\documentclass[letterpaper, 10 pt, journal, twoside]{IEEEtran}
\usepackage{amsmath,amsfonts}
\usepackage{algorithmic}
\usepackage{algorithm}
\usepackage{array}
\usepackage[caption=false,font=normalsize,labelfont=sf,textfont=sf]{subfig}
\usepackage{textcomp}
\usepackage{stfloats}
\usepackage{url}
\usepackage{verbatim}
\usepackage{graphicx}
\usepackage{cite}
\usepackage{booktabs}
\usepackage{tabularx}
\usepackage{multirow}
\usepackage{makecell}
\begin{document}

\title{Mastering Diverse, Unknown, and Cluttered Tracks for Robust Vision-Based Drone Racing}

\author{Feng Yu$^*$, Yu Hu$^*$, Yang Su, Yang Deng, Linzuo Zhang, and Danping Zou
\thanks{Manuscript received: August 15, 2025; Revised: October 12, 2025; Accepted: November 28, 2025.}
\thanks{This paper was recommended for publication by Editor G. Loianno upon evaluation of the Associate Editor and Reviewers' comments. This work was supported
by National Key R\&D Program of China (2022YFB3903802) and National
Science Foundation of China (62073214). \textit{(Corresponding author: Danping Zou.)}} 
\thanks{$^*$ These authors contributed equally to this work.}
\thanks{The authors are with the Shanghai Jiao Tong University, Shanghai 200240,
China (e-mail: dpzou@sjtu.edu.cn).}%
\thanks{Digital Object Identifier (DOI): see top of this page.}}

\markboth{IEEE Robotics and Automation Letters. Preprint Version. NOVEMBER, 2025}
{Yu \MakeLowercase{\textit{et al.}}: Mastering Diverse, Unknown, and Cluttered Tracks
for Robust Vision-Based Drone Racing}

\IEEEpubid{0000--0000/00\$00.00~\copyright~2021 IEEE}

\maketitle

\begin{abstract}
Most reinforcement learning (RL)-based methods for drone racing target fixed, obstacle-free tracks, leaving the generalization to unknown, cluttered environments largely unaddressed.  This challenge stems from the need to balance racing speed and collision avoidance, limited feasible space causing policy exploration trapped in local optima during training, and perceptual ambiguity between gates and obstacles in depth maps-especially when gate positions are only coarsely specified.
To overcome these issues, we propose a two-phase learning framework: an initial soft-collision training phase that preserves policy exploration for high-speed flight, followed by a hard-collision refinement phase that enforces robust obstacle avoidance. An adaptive, noise-augmented curriculum with an asymmetric actor-critic architecture gradually shifts the policy’s reliance from privileged gate-state information to depth-based visual input. We further impose Lipschitz constraints and integrate a track-primitive generator to enhance motion stability and cross-environment generalization.
We evaluate our framework through extensive simulation and ablation studies, and validate it in real-world experiments on a computationally constrained quadrotor. The system achieves agile flight while remaining robust to gate-position errors, 
developing a generalizable drone racing framework with the capability to operate in diverse, partially unknown and cluttered environments.
\end{abstract}


\section{Introduction}
\IEEEPARstart{A}{utonomous} drone racing represents a significant challenge in robotics, inspired by human first-person-view (FPV) racing competitions that require  precise control and split-second decision-making. The task requires high-speed, collision-free navigation through a sequence of gates with minimal lap time, often aiming to surpass the performance of expert human pilots \cite{hanover2024autonomous,kaufmann2023champion}. Consequently, it has become a benchmark problem for evaluating the agility, environmental adaptability, and real-time decision-making capabilities of autonomous drones.

\begin{figure}
\centering
\includegraphics[width=\linewidth]{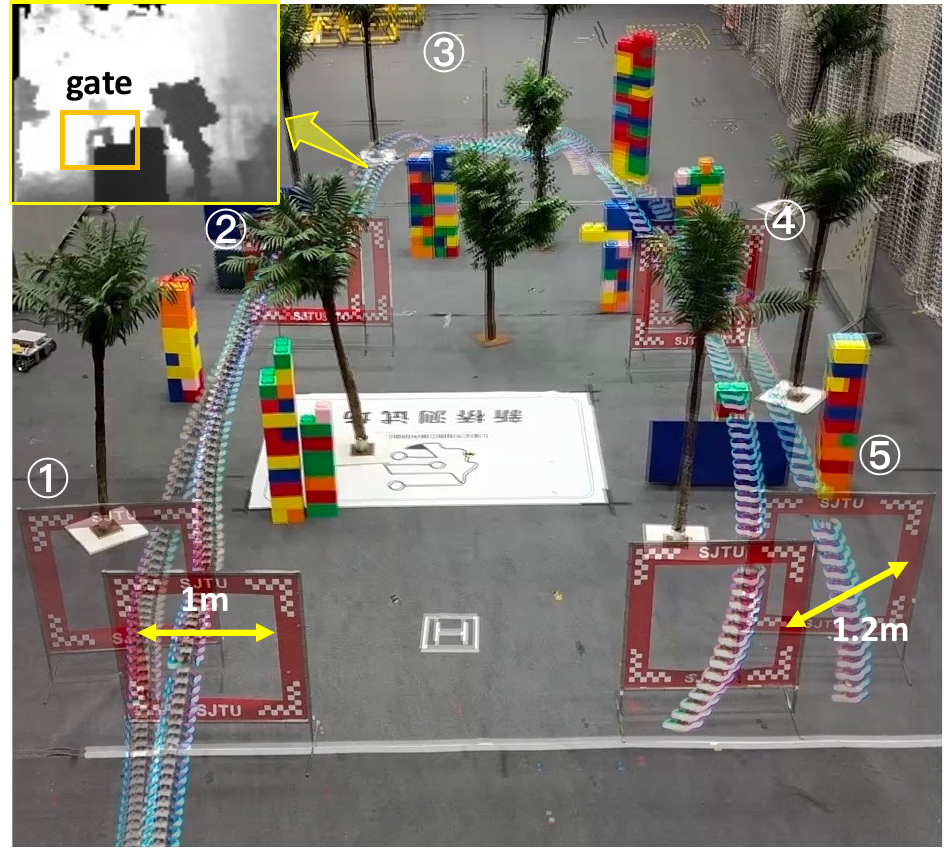}
\caption{
Given corse prior information about gate positions and their order, our trained policy enables the quadrotor to safely traverse the unseen track with random obstacles using only onboard depth sensing and computation, demonstrating strong robustness to gate-position errors and obstacles. 
}
\label{u-ratck-move}
\end{figure}

In recent years, a number of competitions have been launched to foster accelerated progress within the field, such as the IROS’16-19 Autonomous Drone Racing series \cite{10.1007/s11370-018-00271-6}, NeurIPS 2019’s Game of Drones \cite{madaan2020airsim}, and the 2019 AlphaPilot Challenge \cite{foehn2022alphapilot}. In these competitions, the methodology has progressively shifted from traditional perception\textendash planning\textendash control pipelines to end-to-end reinforcement learning. While conventional approaches suffer from cumulative errors and processing latency that limit maximum achievable speeds \cite{HowFast}, RL-based methods significantly reduce inference time and enable direct optimization of flight performance \cite{song2023reaching}—potentially surpassing the capabilities of human pilots. However, existing RL solutions typically require extensive training on fixed, obstacle-free tracks, resulting in poor generalization and limited safety. This raises a critical research question: \textit{Can we train a single depth-vision-based policy that generalizes across diverse, unknown tracks, adapts to noise in gate position commands, and maintains aggressive gate traversal in cluttered environments?}
\IEEEpubidadjcol

Compared with standard drone racing, which is typically formalized as a time-optimal control problem in a fixed environment that pushes vehicle's physical limits, racing in unknown, cluttered environments introduces an additional requirement: the policy must jointly optimize for speed, safety, and generalization to novel scenarios. This creates an inherent trade-off between optimality and adaptability, giving rise to several key challenges. 
First, an exploration-exploitation conflict arises under high-speed, collision-prone settings: high-speed racing favors exploiting in-track progress, whereas obstacle-avoidance necessitates exploring collision-free spaces. Early termination due to collisions exacerbates this imbalance, as early terminated episodes offer limited learning signals for policy improvement.
Second, gates and obstacles often share similar depth cues, such as thin structures against cluttered backgrounds, yet require opposed control responses: aggressive approach versus cautious avoidance. This perceptual ambiguity hinders the policy’s ability to acquire discriminative representations without explicit semantic supervision, reducing adaptability to unfamiliar scenarios. 
Finally, the diversity of track layouts and obstacle configurations imposes stricter demands on the robustness and generalization capability of policy.

\textbf{Contribution}: This work proposes a novel RL framework for agile and robust vision-based racing. We introduce a two-phase training method that explicitly models rigid-body collisions in different manners to enhance both racing speed and obstacle avoidance.
To distinguish between gates and obstacles in the depth map, we integrate an adaptive noise-augmented curriculum within an asymmetric actor-critic training framework. 
Simultaneously, the track primitive generator composes diverse layouts to enhance policy generalization, while Lipschitz constraints promote motion smoothness for safe flight.
To accelerate convergence and reduce the sim-to-real gap, we develop DiffLab \footnote{https://github.com/MasterDroneRacing/MasterRacing}, a highly parallelized and high-fidelity drone simulator built on NVIDIA Isaac Lab \cite{mittal2023orbit}.
Finally, we validate the efficacy of our framework through extensive simulations and real-world flight experiments, demonstrating robust real-world performance as shown in Fig. \ref{u-ratck-move}. 
To the best of our knowledge, this is the first vision-based  autonomous drone racing system operating in diverse, unknown, and cluttered real-world environments.

\section{Related Work}
\subsection{Autonomous Drone Racing without Obstacles}
Optimization-based methods are predominantly designed as state-based approaches. They employ rigorous mathematical formulations to derive explicit solutions through constrained optimization frameworks. The problem is typically formulated as a minimum-time trajectory planning task with discrete waypoints constraints in obstacle-free scenarios. Foehn et al. \cite{foehn2021time} introduces a trajectory progression parameterization scheme, enabling concurrent optimization of temporal allocation and spatial trajectory design. However, this approach is computationally expensive and subsequent research attempts to address this by Model Predictive Contouring Control (MPCC) \cite{romero2022model}, which achieves real-time implementability and simultaneously maximizes path progression while minimizes tracking deviation, and its extended research \cite{romero2022time} demonstrates inherent robustness against environmental variations and external perturbations. However, these state-based methods heavily rely on the drone and gate states without incorporating visual perception systems, thereby limiting environmental understanding and resulting in failures in obstacle-dense environments.

RL-based methods have been shown to achieve high-speed flight \cite{nagami2021hjb}. For example, in \cite{song2021autonomous}, an end-to-end racing network is trained in a simulator and the higher-level commands output by the network are used as tracking command for the MPC controller during real flight. Although the learned policy results in slower performance than optimization based methods, it shows adaptability to gate pose variations and generalization to unseen tracks to a certain extent. The state-based policy can also serve as a teacher network to distill vision-based policies \cite{fu2023learning, xing2024bootstrapping}. Notably, Kaufmann et al. \cite{kaufmann2023champion} accurately model closed-loop systems by combining abstract representations and learned residual models, enabling the trained Swift system to outperform human pilots in real-world scenarios. Furthermore, to improve controller optimality, Ferede et al. \cite{ferede2024end} proposes a high-speed controller using end-to-end RL to directly output motor commands. In addition, Wang et al. \cite{wang2024environment} introduces an environment-shaping framework to enhance the generalization of RL agents across diverse and unseen tracks without retraining.

\subsection{Autonomous Drone Racing in Cluttered Environments}
Several optimization- and planning-based methods have demonstrated effectiveness in obstacle-rich scenarios. Penicka et al. \cite{penicka2022minimum} propose a sampling-based method with incrementally complex quadrotor modes to compute time-optimal trajectories, but it lacks generalization to novel environments. The winning solution of the 2022 DJI RMUA UAV Challenge \cite{wang2023polynomial} proposes an online replanning framework for handling dynamic gates, though its trajectory-based approach depends on accurate state estimates of both the gates and the drone.

RL-based approaches remain relatively underexplored in obstacle-aware drone racing. Penicka et al. \cite{penicka2022learning} address this challenging task by using a topological path planner to guide learning the state-based racing policy. However, their policy is trained on fixed environments and does not generalize to unseen tracks. Other recent RL methods \cite{liu2024learning, xiao2024time} apply advanced techniques but lack real-world validation. In contrast, our approach trains a vision-based policy capable of zero-shot sim-to-real transfer, enabling autonomous drone racing in unseen, cluttered real-world environments.

\section{Methodology}
\begin{figure*}[t]
\centering
\includegraphics[width=1.0\textwidth]{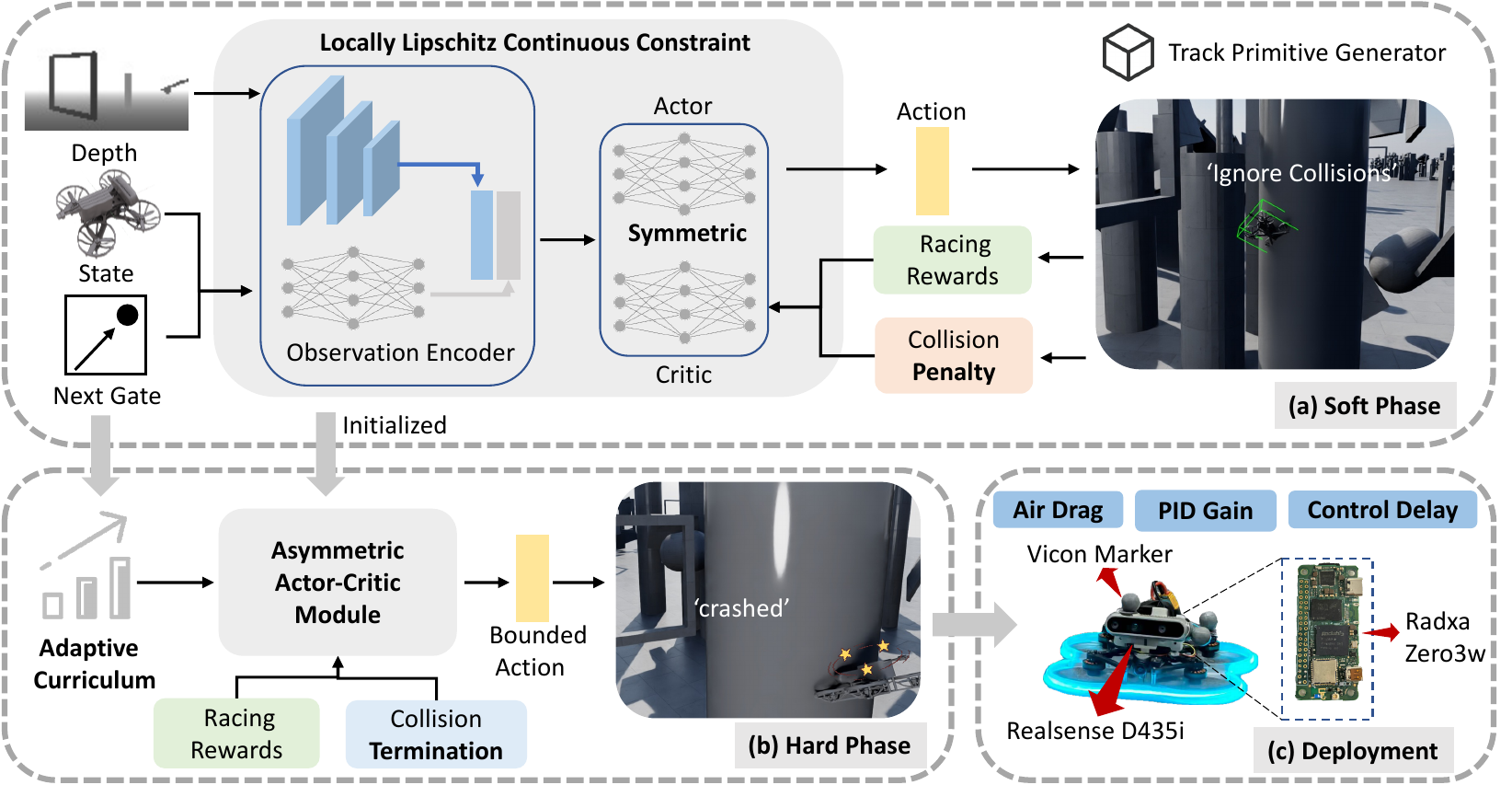}
\caption{The proposed two-phase framework for quadrotor racing comprises: (a) \textbf{Soft Collision Phase}: Depth observations, next gate position commands and drone states are encoded into shared embeddings for a locally Lipschitz constrained actor-critic network. Actions are executed in the collision-free simulator generated by the predefined track primitive generator to encourage racing. (b) \textbf{Hard Collision Phase}: The pre-trained policy is fine-tuned with curriculum noise injected into next gate position commands, while interacting with a rigid-body simulator enforcing real collision effects. (c) \textbf{Deployment}: After system identification aligning simulation and real dynamics, the policy is deployed on a physical quadrotor using Intel RealSense D435i for depth perception and VICON for state estimation.}
\label{framework}
\end{figure*}

\subsection{Overview}
In this work, we tackle vision-based drone racing in diverse, unknown, and cluttered environments, where the drone navigates using noisy gate position commands and perceives the surroundings solely through depth data. This task is highly challenging due to limited field of view, ambiguous perception from depth-only sensing, command inaccuracies from coarse gate positions, and environmental noise.
As illustrated in Fig. \ref{framework}, our framework consists of three key components. First, we decompose the learning process into two progressive optimization stages (Sec. \ref{two_stage}). Second, we propose an adaptive noise-augmented curriculum learning scheme combined with an asymmetric actor–critic architecture, integrated with a track primitive generator (Sec. \ref{learn_behavior}). Finally, we introduce DiffLab, a high-fidelity, parallelized simulator (Sec. \ref{sim_and_dynamics}). Implementation details, including system identification, domain randomization, and Lipschitz regularization are provided in Sec. \ref{training_details}.

\subsection{Learning Obstacle-Aware Agile Racing}
\label{two_stage}
Reinforcement learning for racing in complex environments frequently encounters early termination due to rich collisions, which hinders comprehensive exploration of trial-and-error policies and breaks balance between speed optimization and obstacle avoidance. Inspired by \cite{zhuang2023robot, wang2025beamdojo}, we develop a two-phase reinforcement learning framework for racing which composes of staged soft or hard collision constraints. With the soft constraints, the agent can execute long trajectories to learn aggressive racing behaviors. Afterwards, it will be fine-tuned under the hard constraints to develop better avoidance ability.
\begin{align}
{r}_{soft\_collision}&=-\sum_{p}\mathbb{I}[p], \label{collision_r_1} \\
{r}_{hard\_collision}&=
\left\{
\begin{aligned}
-&1, \text{if collided}\\
& 0, \text{else}.
\end{aligned}
\right.
\label{collision_r_2}
\end{align}

\begin{figure}
\centering
\includegraphics[width=\linewidth]{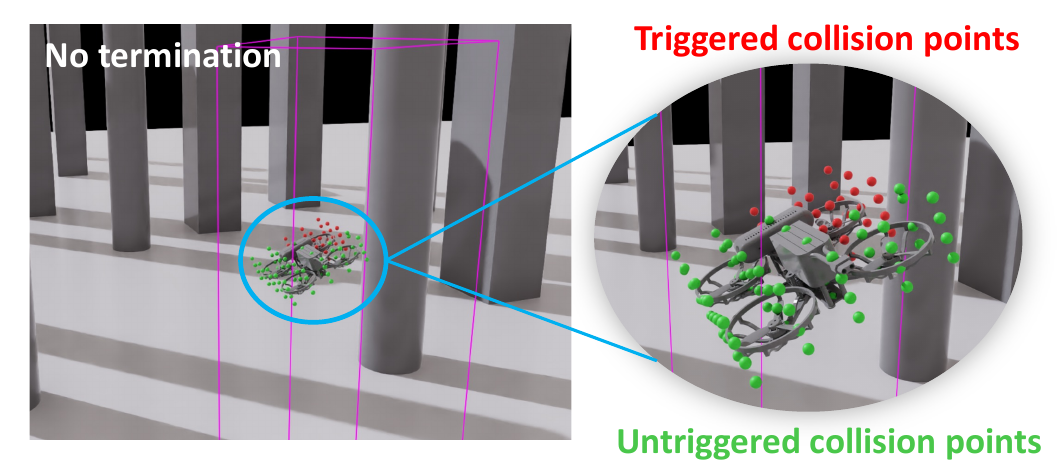}
\caption{The drone completes the first phase without collision termination. Collision points are uniformly placed on the collider box (green/red lattices), and a mild penalty is computed based on activated points (red).}
\label{collision_points}
\end{figure}

\textbf{Soft-Collision Phase}.  To improve racing capability, the drone is trained without rigid body collision effects in this phase, enabling it to pass through obstacles without disrupting flight or triggering early termination. This preserves trajectory continuity, facilitating more complete scene coverage for positive sample collection and racing optimization. The remaining scene configuration remains consistent with real-world conditions. To expose the drone to realistic dynamics and prevent it from over-prioritizing speed over safety, we introduce a mild collision penalty with small weight, allowing the policy to begin balancing racing and obstacle avoidance. As illustrated in Fig. \ref{collision_points}, the penalty is computed based on the number of collision points penetrating into the environment mesh, as described in \eqref{collision_r_1}, where $\mathbb{I}[p] \in [0, 1]$ indicates whether the point $p$ penetrates the environment mesh. For detection, collision points are uniformly placed on the drone body, and Warp\footnote{https://nvidia.github.io/warp/} is used to check their intersections with the environment mesh in parallel. 

\textbf{Hard-Collision Phase}. We fine-tune the pre-trained policy from the first phase in the physically realistic scenarios, where collisions can occur, triggering early termination and penalty signals as \eqref{collision_r_2}. This forces the drone to adhere to the scene settings and learn safer, more precise behaviors. Therefore, unlike the first phase, which primarily focuses on racing skill acquisition, this phase prioritizes exploration to better balance speed maximization and obstacle avoidance. Consequently, the weights assigned to the smoothness and collision penalties are also increased at this phase. Furthermore, an adaptive noise curriculum (see Sec. \ref{learn_behavior}) is employed in this phase to enhance the policy’s ability to visually recognize gates from depth inputs.

\begin{figure}
    \centering
    \includegraphics[width=\linewidth]{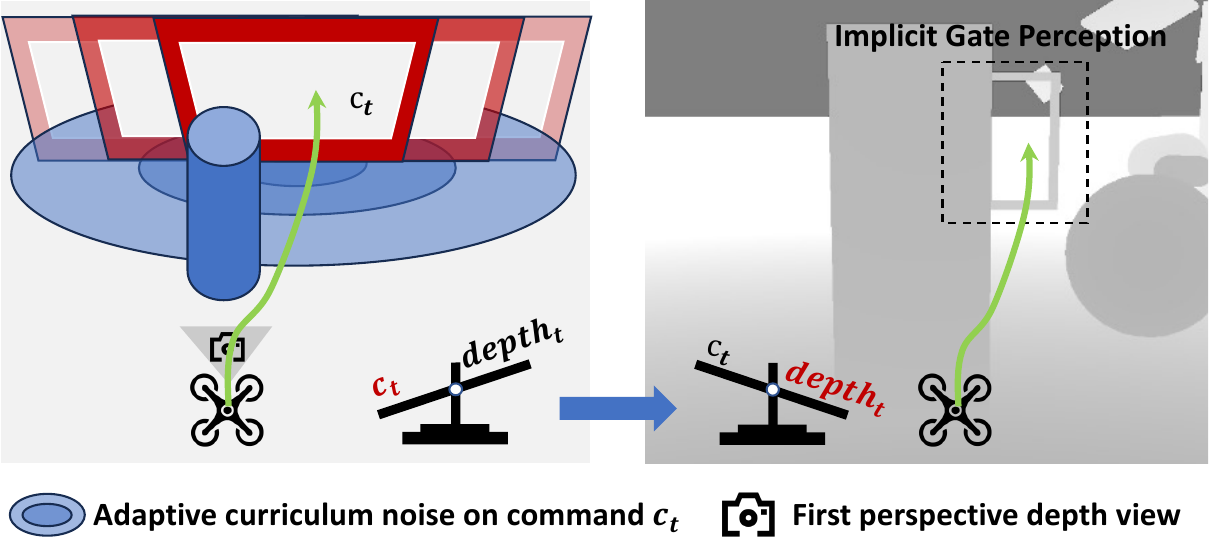}
    \caption{ The policy’s reliance gradually shifts from next gate position commands to depth via the adaptive noise-augmented curriculum and asymmetric actor-critic framework. 
    }
    \label{fig:ada-curr}
\end{figure}

\subsection{Curriculum Learning for Generalizable Racing} \label{learn_behavior}
\textbf{Adaptive Noise-Augmented Curriculum}. Curriculum learning is a widely used technique for acquiring complex behaviors by decomposing difficult tasks into a sequence of progressively harder subtasks \cite{hoeller2024anymal, zhang2024learning}. A common implementation involves gradually increasing the complexity of the environment, such as by raising the density of obstacles to enhance navigation robustness \cite{xu2025navrl}. In this work, beyond applying curriculum learning to environment complexity to enhance obstacle avoidance capability, we further introduce a noise curriculum on the next-gate position commands. 

Specifically, the adaptive noise-augmented curriculum works as follows: (1) increase the noise level if the policy performs well under the current noise, (2) decrease it if performance falls below a predefined tolerance, and (3) keep it unchanged if performance is moderate. The update rule is:
\begin{equation}
^i\mu_{t+1} = ^i\mu_{t} * (1 + \alpha_1)^{\mathbb{I}[n>3]} * (1 - \alpha_2)^{\mathbb{I}[n<3]} \label{curriculum_design}
\end{equation}
Here, $^i\mu_t$ denotes the upper bound of the uniform distribution $\mathcal{U}(-\mu, \mu)$ controlling the single-axis position noise of the $i$-th agent's command  during episode $t$. The indicator $\mathbb{I}[n>3]$ checks whether the agent successfully passes more than three gates, where $n$ represents the number of passed gates. When $n=3$, both indicators in \eqref{curriculum_design} are false, leaving the noise level unchanged and encouraging the policy to better adapt to the current noise level. Here empirical observations indicate that adopting 3 gates as the baseline yields a more rational adjustment rate for curriculum difficulty. $\alpha_1$ and $\alpha_2$ denote the growth and decay rates of the noise level, respectively, with $\alpha_2 > \alpha_1$ so that failed agents are reset with less noisy commands for easier relearning. Because noise adjustment is performance-driven, it avoids the performance degradation that would occur from directly introducing excessive noise from scratch, maintaining a balance between noise intensity and racing performance.

Given the noise introduced by curriculum, we integrate the framework with an asymmetric actor-critic setup \cite{pinto2017asymmetric}. Specifically, as noise gradually increases, these command observations become unreliable, complelling the actor to implicitly extract gate-relevant features from depth rather than noisy command observations for generating reward-maximizing actions, while the critic still uses the ground-truth commands for precise and robust training. 
This combination enables a gradual shift in reliance—from external command guidance to the agent’s own learned perception, thereby promoting policy robustness and enabling flight behaviors that are aware of gate structures in the depth map, as illustrated in Fig. \ref{fig:ada-curr}.

\textbf{Track Primitive Generator}. Although the adaptive noise-augmented curriculum initially facilitates the policy’s adaptation to various gate position inputs, further adaptation to diverse track configurations remains essential. Based on prior observations of typical motion primitives in drone racing tracks, we design three types of training tracks: circular, zigzag, and elliptical. The circular track includes both clockwise and counterclockwise directions, implicitly capturing left and right turning behaviors. The zigzag track extends this by incorporating straight-line flight. The elliptical track integrates the motion patterns of both circular and zigzag tracks and further requires autonomous switching between them. Consequently, the policies trained on these three track types acquire transferable flight proficiencies—including straight-line flight, left turns, and right turns—that enable effective racing on most tracks.

\subsection{Policy Training} \label{training_details}
\subsubsection{Observation}
The policy observations $\boldsymbol{o}_{t}$ at time $t$ consists of: 
\begin{equation}
    \boldsymbol{o}_t=[^b\boldsymbol v_t, \mathbf{r_3}, ^b\boldsymbol{\delta p}_{t}^1, ^b\boldsymbol{\delta p}_t^2,\boldsymbol{a}_{t-1}, \boldsymbol{I}_{t}]
\end{equation}
Here $^b\boldsymbol v_t$ denotes the linear base velocity in the body frame, and $\mathbf{r}_3$ represents the third row of the rotation matrix, encoding roll and pitch angles. The command $^b\boldsymbol{\delta p}_{t}^1$ and $^b\boldsymbol{\delta p}_t^2$ represent the relative translation from the drone to the next gate, and from the next gate to the second subsequent gate, respectively, both expressed in the body frame. Both translation vectors are injected with noise, as described in ~\ref{learn_behavior}. The previous action $\boldsymbol{a}_{t-1}$ is included to encourage smooth control, making the policy aware of past commands rather than purely reactive. Following \cite{zhang2025learning}, which employs a low-resolution depth image for obstacle avoidance, we downsample the depth image $\boldsymbol{I}_{t}$ to  $96 \times 72$. This reduces observation dimensionality, improving sample efficiency,  and retaining sufficient gate-relevant information.

\subsubsection{Action}
The policy outputs a control command at each timestep $t$, comprising mass-normalized collective thrust and body rates (CTBR) following \cite{kaufmann2022benchmark}.

\subsubsection{Reward}
Most reward functions follow \cite{kaufmann2023champion}, with minor modifications for the progress and perception rewards. The progress reward directs the drone toward the target, while the perception reward encourages it to face the target, enhancing safety and task success. The reward function is largely consistent across the two-phase training, with only slight adjustments to component weights (Table \ref{reward_functions}).
\subsubsection{Network Design}
Our network architecture is shown in Fig. \ref{framework}. An MLP is used to process the state-only and command observations, while a CNN processes the depth images. The resulting feature vectors are concatenated and fed into subsequent MLP layers, which finally output CTBR control commands or value estimates. 
\subsubsection{Gate Crossing Detector}
Because noise in the relative gate positions makes position-based gate-crossing checks unreliable, we design an automatic gate-crossing detector based on network embeddings. Specifically, we collect a balanced set of positive and negative samples in the simulator using a well-trained policy. Each sample consists of a pair: network embeddings and a corresponding gate-crossing signal. The prediction head is then trained offline using those samples with the rest of the policy network frozen. This head remains excluded from updates during policy training.

\subsubsection{Sim-to-real Transfer}
To narrow the sim-to-real gap and enhance policy robustness, we apply extensive domain randomization to key drone parameters, including air drag coefficients, mass, inertia, and the PID gains of the angular velocity controller. In addition to the adaptive command noise, Gaussian noise is injected into other observation channels.  Following \cite{yin2025taco, zhang2024robust, kobayashi2022l2c2}, we adopt local Lipschitz continuity constraint (L2C2) to suppress action oscillations.  
This method regularizes both the actor and critic to enforce local Lipschitz continuity, enhancing temporal smoothness of the action and value functions. Specifically, an additional loss term is introduced to bound the output variation between consecutive states:
\begin{align}
    \mathcal{L}_{L2C2} &= \lambda_1 D(\pi_{\theta}(o_t), \pi_{\theta}(\hat{o_t})) + \lambda_2 \Vert V_{\phi}(o_t)-V_{\phi}(\hat{o_t})\Vert ^2_2, \label{l2c2} \\
    \hat{o_t}&=o_t+(o_{t+1}-o_t) \cdot u, u \sim U(-1,1)
\end{align}
where  $\lambda_1$ and $\lambda_2$ are weighting coefficients, $D$ represents the squared Hellinger distance, $\pi_{\theta}$ and $V_{\phi}$ are the policy and value networks parameterized by $\theta$ and $\phi$, respectively, and $\hat{o_t}$ is the interpolated observations between $o_t$ and $o_{t+1}$.

\begin{table}[t]
    \renewcommand\arraystretch{1.5}
    \setlength{\abovecaptionskip}{0pt}
	\caption{Reward Function Terms}
    \label{reward_functions}
    \centering
    \large
    \resizebox{0.5\textwidth}{!}{
	\begin{tabular}{ccc}
		\toprule 
		\textbf{Reward Term} & \textbf{Expression} & \textbf{Weight} \\ 
		\midrule  
		Towards Reward & $\frac{(\boldsymbol{p}^{gate}_w - \boldsymbol{p}_{wb}) \cdot \boldsymbol{v}_w}{\Vert \boldsymbol{p}^{gate}_w - \boldsymbol{p}_{wb} \Vert \cdot \Vert \boldsymbol{v}_w\Vert }$ &1 \\
        Body Rate Penalty & $\Vert \boldsymbol{\omega}_{b} \Vert$ & $-0.02,-0.1^*$ \\
        Action Rate Penalty & $\Vert \boldsymbol{a}_t - \boldsymbol{a}_{t-1}\Vert ^2$ & $-0.01,-0.05^*$ \\
        Collision Penalty & Eq. \eqref{collision_r_1} & 50 \\
        & Eq. \eqref{collision_r_2}$^*$ & $100^*$ \\
        Perception Reward & \makecell{$Normalize(\boldsymbol{q}_t^{-1} \odot (\boldsymbol{p}_{w}^{gate} - \boldsymbol{p}_{wb}))$ \\ $\cdot [1, 0,0]^T $} & 0.1 \\
        Success Reward & \makecell{$\mathbb{I}[\Vert \boldsymbol{p}_w^{gate} - \boldsymbol{p}_{wb} \Vert < 0.35] $ \\  $\cdot \frac{1}{1+\Vert \boldsymbol{p}_w^{gate} - \boldsymbol{p}_{wb} \Vert^2}$} & $10, 20^*$ \\
        Bad Pose Penalty$^*$ & $\mathbb{I}[\Vert roll \Vert>\frac{\pi}{2}]\cdot \mathbb{I}[\Vert pitch \Vert>\frac{\pi}{2}]^*$ & $-30^*$ \\
		\bottomrule 
        \multicolumn{3}{l}{$*$ denotes the settings for the second phase.}
	\end{tabular}
    }
\end{table}

\subsection{Quadrotor Simulator} \label{sim_and_dynamics}

According to quadrotor dynamics and widely-used control methods \cite{faessler2017differential,lee2010control} , we develop DiffLab based on Isaac Lab. Compared with previous work, we focus more on the sim-to-real gap. Specifically, we first adopt a more accurate air drag model that includes both first-order and second-order terms. We then account for control delay to better capture the angular velocity response. All parameters are calibrated using real flight data. Similar to AirGym \cite{huang2025general}, we provide multiple control levels, including position commands and yaw (PY), linear velocity commands and yaw (LV), CTBR and single-rotor thrusts (SRT). These controllers operate in a cascade PID loop, and the results are ultimately converted into total thrust and torque acting on the rigid body of the quadrotor.

\section{Experiements}
We employ the Proximal Policy Optimization (PPO) algorithm \cite{schulman2017proximal} to train the policy in DiffLab. Training on a standard computer with an Intel i7 CPU and an Nvidia RTX 4090 GPU, the policy converges in approximately 4 hours.

\begin{figure}
    \centering
    \includegraphics[width=\linewidth]{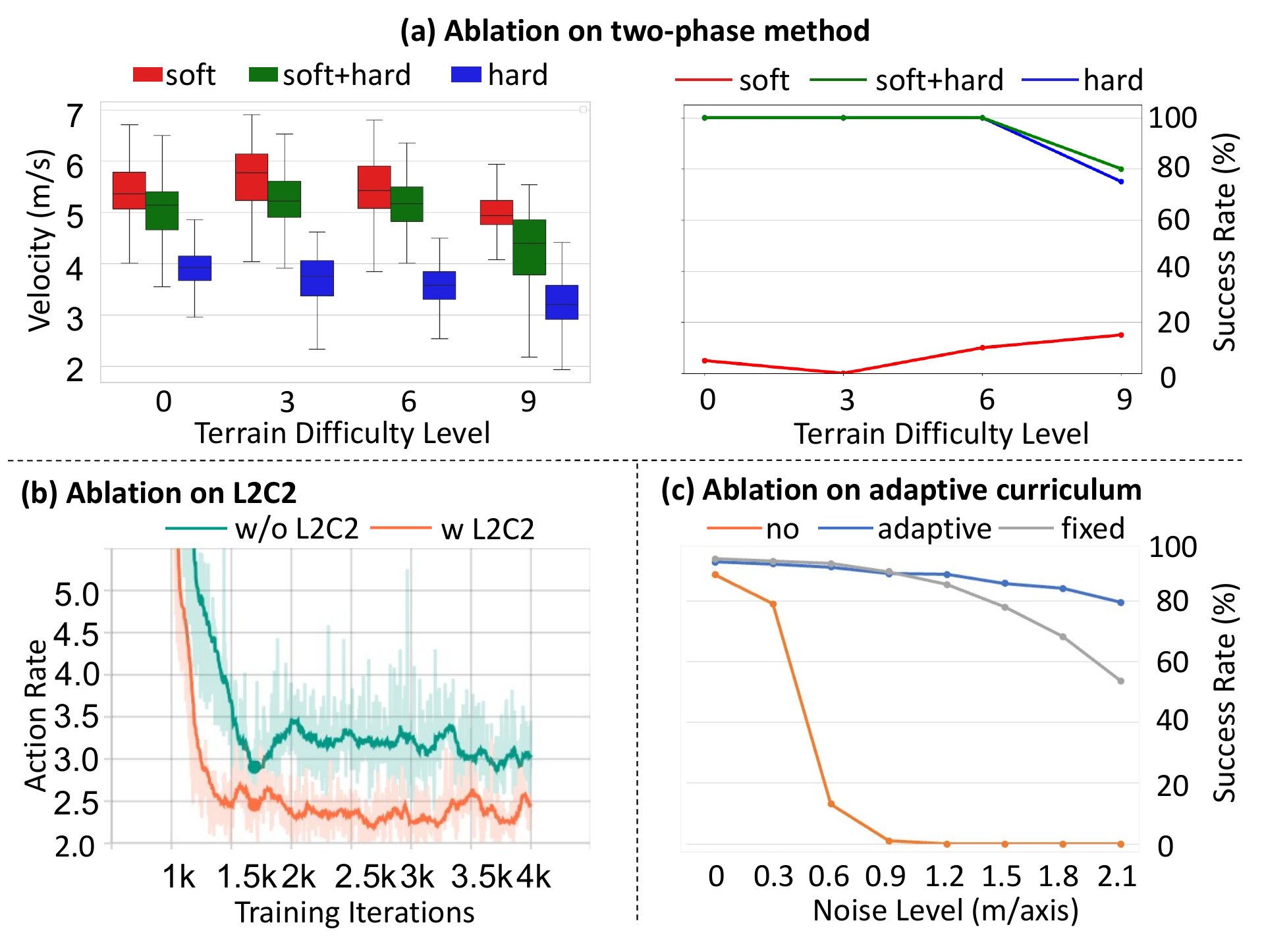}
    \caption{\textbf{Ablation results.} (a) Comparison of velocity and success rate across training phases: soft-collision only (red), combined soft- and hard-collision (green), and the baseline hard-collision-only method (blue). (b) Action rate curves showing policy behavior with and without L2C2 regularization. (c) Success rate comparison under three noise curricula: no noise (orange), fixed noise (gray), and our adaptive noise curriculum (blue).}
    \label{ablation_results}
\end{figure}

\subsection{Ablation Studies} \label{ablation_exp}
This section provides an analysis of the design of each module of our algorithm. We measure the effectiveness using three metrics: average speed, success rate, and action rate, where success rate refers to the proportion of 20 drones completing the track without collision within 8 seconds.
\begin{itemize}
    \item \textbf{Two-Phase vs One-Phase}: We compare the performances of policies trained via two-phase and one-phase methods, with all settings maintained consistently between these approaches. Fig. \ref{ablation_results} (a) illustrates that the two-phase method achieves higher average speed across terrains of all difficulty levels\footnote{Difficulty is determined by obstacle density, track layout, and gate size; a few obstacles remain even at level 0.}, while its second phase maintains the similar success rate as the one-phase method. These results demonstrate that the two-phase method can exploit racing ability gained in the first phase while exploring and improving obstacle avoidance capability in the second phase. In contrast, the one-phase policy is more conservative and unable to fully exploit its racing potential.
    \item  \textbf{Adaptive Noise-Augmented Curriculum}: We validate the effectiveness of our adaptive noise-augmented curriculum via comparison with noise-free and fixed-noise (1 m/axis) baselines. Results in Fig. \ref{ablation_results} (c) demonstrate that the curriculum significantly enhances the policy robustness under various level of noise. For the noise-free command policy, performance degrades notably as noise increases, with success rate approaching 0 at 0.9 m noise. Similarly, the fixed-noise policy shows severe degradation under high noise, achieving only 54\% success rate at 2.1 m noise. In contrast, our method achieves nearly 80\% success rate under the same conditions.
    \item \textbf{Locally Lipschitz Continuous Constraint}: We compare the action rates of the policies trained with and without L2C2, as shown in the Fig. \ref{ablation_results} (b). It can be seen that using L2C2 can significantly reduce the action rate, decreasing motion oscillation by 25\% and resulting in smoother actions.

\end{itemize}

\begin{figure}
    \centering
    \includegraphics[width=\linewidth]{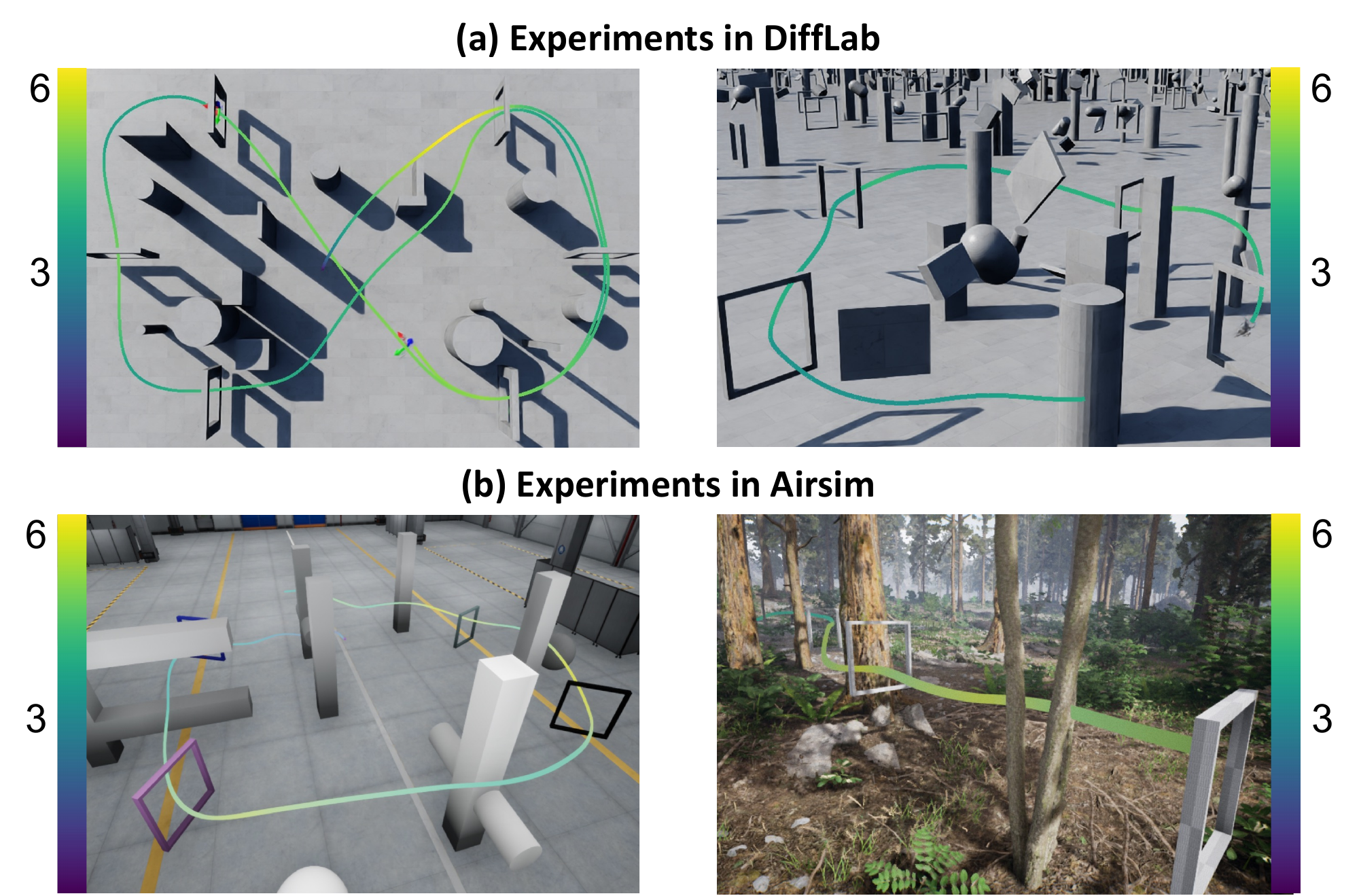}
    \caption{\textbf{Simulation results}: We evaluate the policy in both DiffLab and Airsim simulator on different tracks with different obstacle layouts.}
    \label{sim2sim_results}
\end{figure}

\begin{table}[h!]
\renewcommand\arraystretch{1.4}
\caption{Sim-to-sim Performance}
\label{sim_performance}
\centering
\begin{tabularx}{\linewidth}{c|>{\centering\arraybackslash}X>{\centering\arraybackslash}X|>{\centering\arraybackslash}X>{\centering\arraybackslash}X}
\toprule
\multirow{2}{*}{\textbf{Range (m/axis)}} & \multicolumn{2}{c|}{\textbf{Factory}} & \multicolumn{2}{c}{\textbf{Forest}} \\
& \textbf{SR} & \makecell{\textbf{Max} \\ \textbf{Vel (m/s)}} & \textbf{SR} & \makecell{\textbf{Max} \\ \textbf{Vel (m/s)}} \\
\midrule
$[0.0, 0.0]$     & 10/10 & 5.1  & 10/10 & 5.3  \\
$[-0.3, 0.3]$    & 10/10 & 4.9  & 10/10 & 5.1  \\
$[-0.6, 0.6]$    & 10/10 & 4.4  & 10/10 & 4.6  \\
$[-0.9, 0.9]$    & 7/10  & 4.4  & 8/10  & 4.6  \\
$[-1.2, 1.2]$    & 5/10  & 4.3  & 8/10  & 4.6  \\
\bottomrule
\multicolumn{5}{l}{\footnotesize SR and Max Vel denote success rate and maximum velocity, respectively.}
\end{tabularx}
\end{table}

\subsection{Simulation Experiments}
We conduct extensive experiments in both our DiffLab simulator and the AirSim simulator \cite{airsim2017fsr}. In DiffLab, we employ a structured environment similar to the training setup, while in AirSim, we evaluate the policy in factory and forest environments. As illustrated in Fig. \ref{sim2sim_results}, multi-perspective visualizations of the flight process demonstrate that our policy generalizes effectiely to unseen, realistic environments and novel track layouts, achieving 
peak speeds at 5 m/s in both forest and factory scenes. This result confirms the effectiveness of the track primitive generator, enhancing the policy’s generalization capability to diverse tracks and scenarios.

We further evaluate the policy’s robustness to gate position noise in simulation to demonstrate its gate perception-aware capability. As shown in Table \ref{sim_performance},the policy achieves a 100\% success rate across all scenarios  when the single-axis position error is below 0.6 m. Even with errors approaching 1.2 m, it maintains a success rate above 50\%. Moreover, the maximum speeds exceed 5 m/s in both  test environments, with only moderate performance degradation as noise increases. These results validate the effectiveness of our adaptive noise-augmented curriculum in enhancing the policy's robustness to gate position errors. Notably, the gradual decline in success rate under high-noise conditions is not due to collisions but rather to incorrect commands caused by excessive noise. These misguiding 'next-gate position' commands lead the drone to deviate significantly from the track and fail to cross gates. While this results in task failure, it demonstrates the policy’s strong obstacle avoidance capabilities.

\begin{figure*}[t]
    \centering
    \includegraphics[width=\linewidth]{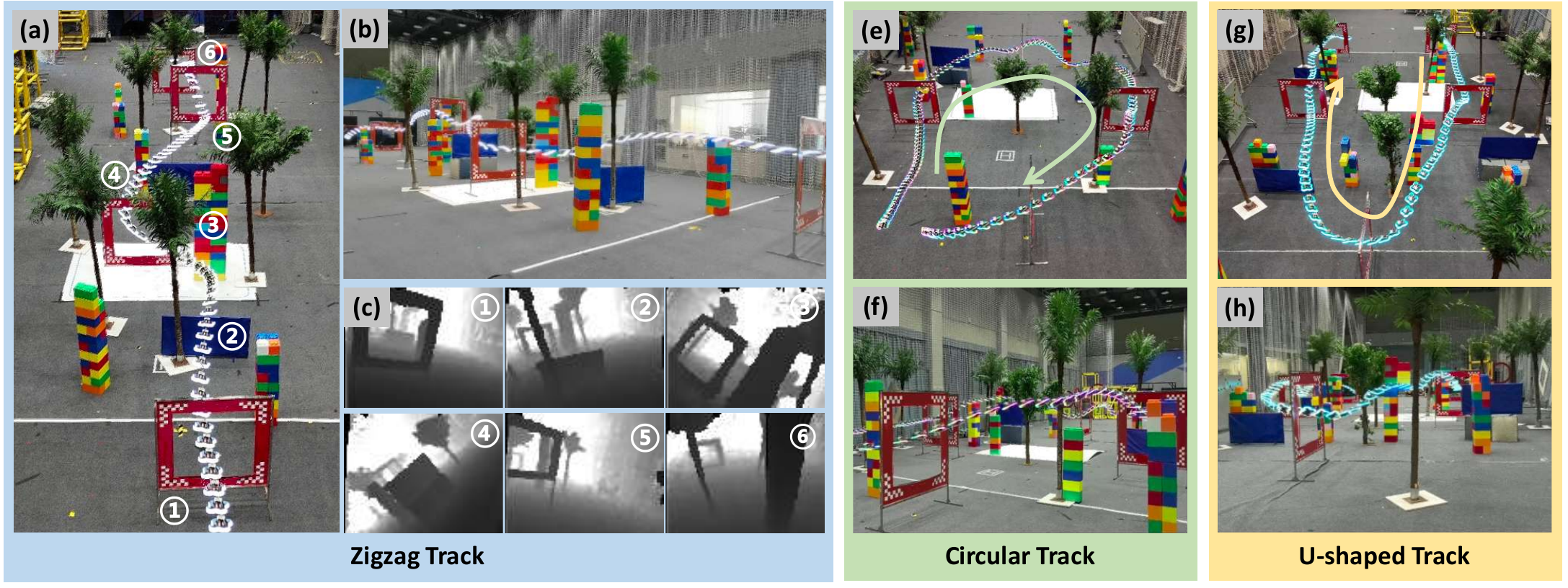}
    \caption{\textbf{Real-world results.} We validate our framework in various unknown, cluttered environments, including (a) a zigzag track, (d) a circular track, and (f) a U-shaped track with randomly placed obstacles such as trees and boxes. The quadrotor's flight trajectory is visualized by overlaying its position in the video footage (please refer to the supplementary video for more details). (b), (e), and (g) show side views of the trajectories, while (c) presents the front depth view corresponding to (a).}
    \label{fig:real-exp}
\end{figure*}

\begin{figure}
    \centering
    \includegraphics[width=\linewidth]{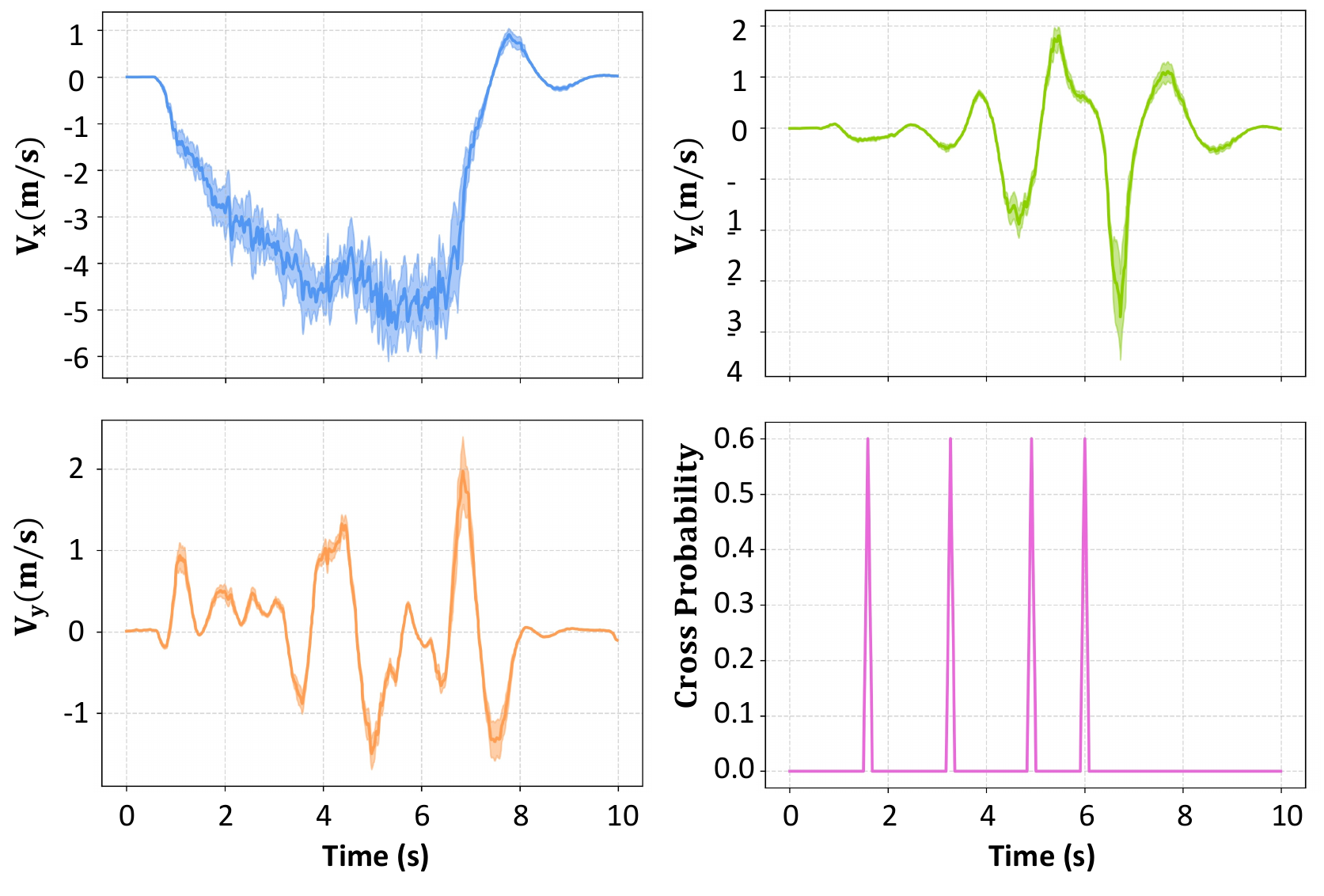}
    \caption{\textbf{Velocity and Gate Detection Results}. We present the velocity measurements along all three axes, recorded by the motion capture system, as well as the gate detection results output by our policy during flight on the zigzag track shown in Fig. \ref{fig:real-exp} (a).}
    \label{fig:real-velocity}
\end{figure}

\subsection{Real World Experiments}
We validate our policy on a lightweight quadrotor platform (0.46 kg), equipped with an Intel Realsense D435i depth camera and a Radxa Zero3W onboard computer (1.6GHz quad-core A55 CPU with 1TOPS NPU)\footnote{https://radxa.com/products/zeros/zero3w} as illustrated in Fig.\ref{framework} (c). A high-precision motion capture system running at 100Hz provides accurate drone motions, while low-level control is executed via a Betaflight controller.

We conduct experiments in various track and obstacle layouts, including a zigzag track, a circular track and a U-shaped track, as shown in Fig. \ref{fig:real-exp}. The quadrotor successfully navigates through all gates in cluttered environments, achieving a maximum speed exceeding 5 m/s, as illustrated in Fig. \ref{fig:real-velocity}. Additionally, accurate gate-crossing detection results are also presented in Fig. \ref{fig:real-velocity}. These results demonstrate the robustness and generalization capabilities of our method in real-world scenarios.

\begin{figure}
    \centering
    \includegraphics[width=0.9\linewidth]{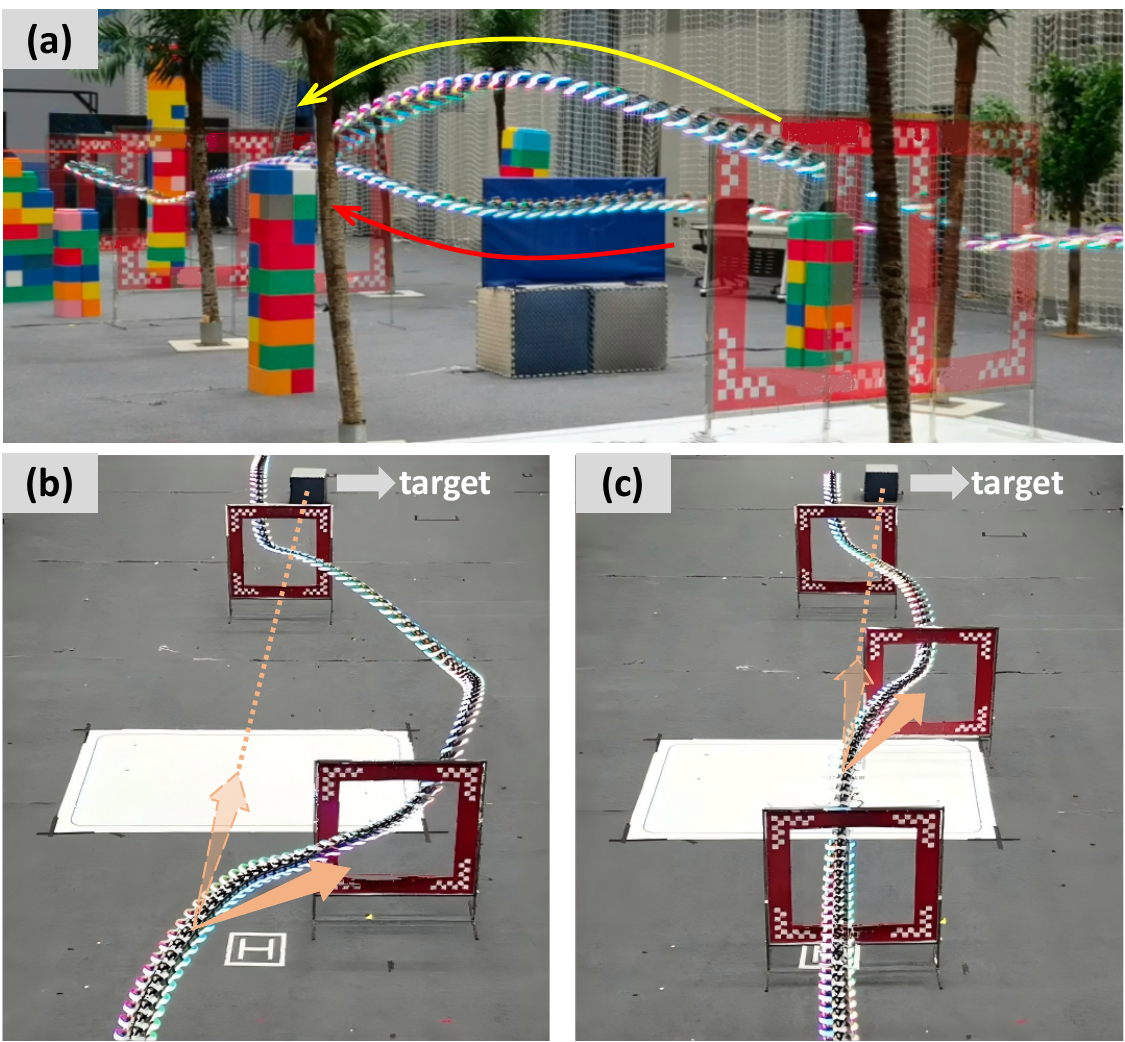}
\caption{\textbf{Robustness Tests.} (a) Gate positions are randomly perturbed while keeping the gate position commands fixed. Trajectories from the perturbed track (second run) are overlaid on the initial flight for comparison. (b, c) The target command is set beyond the track’s end. Despite inaccurate gate commands, the drone demonstrates gate perception-aware behavior by autonomously adjusting its flight to pass through the gates.}
    \label{fig:move-gate}
\end{figure}

To further validate our policy's robustness to gate position errors, we conducted two sets of experiments: one on a zigzag track and the other on a U-shaped track. For each track, we first measured the gate positions and used them as commands for an initial flight. Then, we introduced random perturbations to the gate positions and performed a second flight using the original commands. Results shown in the Fig. \ref{u-ratck-move} and Fig. \ref{fig:move-gate} (a) demonstrate that our policy tolerate noise exceeding 1 meter, flying swiftly and accurately through all gates. This highlights the policy's gate perception-aware capability to correct noisy commands. We further validated this in real-world tests, as shown in Fig. \ref {fig:move-gate} (b) and Fig. \ref{fig:move-gate} (c). The target command is set directly at the end of the track, keeping the gate within the camera's view but misaligned with the command direction. Despite this, the drone adjusted its trajectory to pass through the gate instead of flying straight, confirming the policy’s perception-driven robustness to gate position errors.

\section{Conclusion And Limitation}
In this work, we propose a learning-based approach for vision-based drone racing, enabling generalization from fixed obstacle-free tracks to diverse tracks in cluttered environments. Our method combines two-phase training with an adaptive noise-augmented curriculum to develop robust gate perception-aware racing capability. Extensive experiments validate the effectiveness of the proposed framework.

\textbf{Limitations}: While our approach generalizes well across diverse racing scenarios, there is still room for improvement in the flight speed of drones, and the distribution of scenes will greatly affect the speed of drones. It is possible to further study the balance between optimality and safety in various scenarios, as well as perception-based pose correction to reduce reliance on external localization.

\bibliographystyle{ieeetr}
\bibliography{reference}

@article{hanover2024autonomous,
  title={Autonomous drone racing: A survey},
  author={Hanover, Drew and Loquercio, Antonio and Bauersfeld, Leonard and Romero, Angel and Penicka, Robert and Song, Yunlong and Cioffi, Giovanni and Kaufmann, Elia and Scaramuzza, Davide},
  journal={IEEE Transactions on Robotics},
  year={2024},
  publisher={IEEE}
}

@article{kaufmann2023champion,
  title={Champion-level drone racing using deep reinforcement learning},
  author={Kaufmann, Elia and Bauersfeld, Leonard and Loquercio, Antonio and M{\"u}ller, Matthias and Koltun, Vladlen and Scaramuzza, Davide},
  journal={Nature},
  volume={620},
  number={7976},
  pages={982--987},
  year={2023},
  publisher={Nature Publishing Group UK London}
}

@article{10.1007/s11370-018-00271-6,
author = {Moon, Hyungpil and Martinez-Carranza, Jose and Cieslewski, Titus and Faessler, Matthias and Falanga, Davide and Simovic, Alessandro and Scaramuzza, Davide and Li, Shuo and Ozo, Michael and Wagter, Christophe and Croon, Guido and Hwang, Sunyou and Jung, Sunggoo and Shim, Hyunchul and Kim, Haeryang and Park, Minhyuk and Au, Tsz-Chiu and Kim, Si Jung},
title = {Challenges and implemented technologies used in autonomous drone racing},
year = {2019},
issue_date = {April     2019},
publisher = {Springer-Verlag},
address = {Berlin, Heidelberg},
volume = {12},
number = {2},
issn = {1861-2776},
url = {https://doi.org/10.1007/s11370-018-00271-6},
doi = {10.1007/s11370-018-00271-6},
journal = {Intell. Serv. Robot.},
month = apr,
pages = {137–148},
numpages = {12}
}

@article{foehn2022alphapilot,
  title={Alphapilot: Autonomous drone racing},
  author={Foehn, Philipp and Brescianini, Dario and Kaufmann, Elia and Cieslewski, Titus and Gehrig, Mathias and Muglikar, Manasi and Scaramuzza, Davide},
  journal={Autonomous Robots},
  volume={46},
  number={1},
  pages={307--320},
  year={2022},
  publisher={Springer}
}

@inproceedings{wang2023polynomial,
  title={Polynomial-based online planning for autonomous drone racing in dynamic environments},
  author={Wang, Qianhao and Wang, Dong and Xu, Chao and Gao, Alan and Gao, Fei},
  booktitle={2023 IEEE/RSJ International Conference on Intelligent Robots and Systems (IROS)},
  pages={1078--1085},
  year={2023},
  organization={IEEE}
}

@article{hoeller2024anymal,
  title={Anymal parkour: Learning agile navigation for quadrupedal robots},
  author={Hoeller, David and Rudin, Nikita and Sako, Dhionis and Hutter, Marco},
  journal={Science Robotics},
  volume={9},
  number={88},
  pages={eadi7566},
  year={2024},
  publisher={American Association for the Advancement of Science}
}

@inproceedings{zhang2024learning,
  title={Learning agile locomotion on risky terrains},
  author={Zhang, Chong and Rudin, Nikita and Hoeller, David and Hutter, Marco},
  booktitle={2024 IEEE/RSJ International Conference on Intelligent Robots and Systems (IROS)},
  pages={11864--11871},
  year={2024},
  organization={IEEE}
}

@article{zhuang2023robot,
  title={Robot parkour learning},
  author={Zhuang, Ziwen and Fu, Zipeng and Wang, Jianren and Atkeson, Christopher and Schwertfeger, Soeren and Finn, Chelsea and Zhao, Hang},
  journal={arXiv preprint arXiv:2309.05665},
  year={2023}
}

@inproceedings{song2021autonomous,
  title={Autonomous drone racing with deep reinforcement learning},
  author={Song, Yunlong and Steinweg, Mats and Kaufmann, Elia and Scaramuzza, Davide},
  booktitle={2021 IEEE/RSJ International Conference on Intelligent Robots and Systems (IROS)},
  pages={1205--1212},
  year={2021},
  organization={IEEE}
}

@article{song2023reaching,
  title={Reaching the limit in autonomous racing: Optimal control versus reinforcement learning},
  author={Song, Yunlong and Romero, Angel and M{\"u}ller, Matthias and Koltun, Vladlen and Scaramuzza, Davide},
  journal={Science Robotics},
  volume={8},
  number={82},
  pages={eadg1462},
  year={2023},
  publisher={American Association for the Advancement of Science}
}

@article{penicka2022learning,
  title={Learning minimum-time flight in cluttered environments},
  author={Penicka, Robert and Song, Yunlong and Kaufmann, Elia and Scaramuzza, Davide},
  journal={IEEE Robotics and Automation Letters},
  volume={7},
  number={3},
  pages={7209--7216},
  year={2022},
  publisher={IEEE}
}

@article{penicka2022minimum,
  title={Minimum-time quadrotor waypoint flight in cluttered environments},
  author={Penicka, Robert and Scaramuzza, Davide},
  journal={IEEE Robotics and Automation Letters},
  volume={7},
  number={2},
  pages={5719--5726},
  year={2022},
  publisher={IEEE}
}

@article{liu2024learning,
  title={Learning Generalizable Policy for Obstacle-Aware Autonomous Drone Racing},
  author={Liu, Yueqian},
  journal={arXiv preprint arXiv:2411.04246},
  year={2024}
}

@article{mittal2023orbit,
   author={Mittal, Mayank and Yu, Calvin and Yu, Qinxi and Liu, Jingzhou and Rudin, Nikita and Hoeller, David and Yuan, Jia Lin and Singh, Ritvik and Guo, Yunrong and Mazhar, Hammad and Mandlekar, Ajay and Babich, Buck and State, Gavriel and Hutter, Marco and Garg, Animesh},
   journal={IEEE Robotics and Automation Letters},
   title={Orbit: A Unified Simulation Framework for Interactive Robot Learning Environments},
   year={2023},
   volume={8},
   number={6},
   pages={3740-3747},
   doi={10.1109/LRA.2023.3270034}
}

@article{romero2022model,
  title={Model predictive contouring control for time-optimal quadrotor flight},
  author={Romero, Angel and Sun, Sihao and Foehn, Philipp and Scaramuzza, Davide},
  journal={IEEE Transactions on Robotics},
  volume={38},
  number={6},
  pages={3340--3356},
  year={2022},
  publisher={IEEE}
}

@article{romero2022time,
  title={Time-optimal online replanning for agile quadrotor flight},
  author={Romero, Angel and Penicka, Robert and Scaramuzza, Davide},
  journal={IEEE Robotics and Automation Letters},
  volume={7},
  number={3},
  pages={7730--7737},
  year={2022},
  publisher={IEEE}
}

@article{foehn2021time,
  title={Time-optimal planning for quadrotor waypoint flight},
  author={Foehn, Philipp and Romero, Angel and Scaramuzza, Davide},
  journal={Science robotics},
  volume={6},
  number={56},
  pages={eabh1221},
  year={2021},
  publisher={American Association for the Advancement of Science}
}

@inproceedings{nagami2021hjb,
  title={HJB-RL: Initializing Reinforcement Learning with Optimal Control Policies Applied to Autonomous Drone Racing.},
  author={Nagami, Keiko and Schwager, Mac},
  booktitle={Robotics: science and systems},
  pages={1--9},
  year={2021}
}

@inproceedings{fu2023learning,
  title={Learning deep sensorimotor policies for vision-based autonomous drone racing},
  author={Fu, Jiawei and Song, Yunlong and Wu, Yan and Yu, Fisher and Scaramuzza, Davide},
  booktitle={2023 IEEE/RSJ International Conference on Intelligent Robots and Systems (IROS)},
  pages={5243--5250},
  year={2023},
  organization={IEEE}
}

@article{xing2024bootstrapping,
  title={Bootstrapping reinforcement learning with imitation for vision-based agile flight},
  author={Xing, Jiaxu and Romero, Angel and Bauersfeld, Leonard and Scaramuzza, Davide},
  journal={arXiv preprint arXiv:2403.12203},
  year={2024}
}

@inproceedings{ferede2024end,
  title={End-to-end reinforcement learning for time-optimal quadcopter flight},
  author={Ferede, Robin and De Wagter, Christophe and Izzo, Dario and De Croon, Guido CHE},
  booktitle={2024 IEEE International Conference on Robotics and Automation (ICRA)},
  pages={6172--6177},
  year={2024},
  organization={IEEE}
}

@article{wang2024environment,
  title={Environment as policy: Learning to race in unseen tracks},
  author={Wang, Hongze and Xing, Jiaxu and Messikommer, Nico and Scaramuzza, Davide},
  journal={arXiv preprint arXiv:2410.22308},
  year={2024}
}

@article{wang2025beamdojo,
  title={Beamdojo: Learning agile humanoid locomotion on sparse footholds},
  author={Wang, Huayi and Wang, Zirui and Ren, Junli and Ben, Qingwei and Huang, Tao and Zhang, Weinan and Pang, Jiangmiao},
  journal={arXiv preprint arXiv:2502.10363},
  year={2025}
}

@inproceedings{kaufmann2022benchmark,
  title={A benchmark comparison of learned control policies for agile quadrotor flight},
  author={Kaufmann, Elia and Bauersfeld, Leonard and Scaramuzza, Davide},
  booktitle={2022 International Conference on Robotics and Automation (ICRA)},
  pages={10504--10510},
  year={2022},
  organization={IEEE}
}

@article{xiao2024time,
  title={Time-optimal Flight in Cluttered Environments via Safe Reinforcement Learning},
  author={Xiao, Wei and Feng, Zhaohan and Zhou, Ziyu and Sun, Jian and Wang, Gang and Chen, Jie},
  journal={arXiv preprint arXiv:2406.19646},
  year={2024}
}

@inproceedings{kobayashi2022l2c2,
  title={L2c2: Locally lipschitz continuous constraint towards stable and smooth reinforcement learning},
  author={Kobayashi, Taisuke},
  booktitle={2022 IEEE/RSJ International Conference on Intelligent Robots and Systems (IROS)},
  pages={4032--4039},
  year={2022},
  organization={IEEE}
}

@article{huang2025general,
  title={A General Infrastructure and Workflow for Quadrotor Deep Reinforcement Learning and Reality Deployment},
  author={Huang, Kangyao and Wang, Hao and Luo, Yu and Chen, Jingyu and Chen, Jintao and Zhang, Xiangkui and Ji, Xiangyang and Liu, Huaping},
  journal={arXiv preprint arXiv:2504.15129},
  year={2025}
}

@article{xu2025navrl,
  title={Navrl: Learning safe flight in dynamic environments},
  author={Xu, Zhefan and Han, Xinming and Shen, Haoyu and Jin, Hanyu and Shimada, Kenji},
  journal={IEEE Robotics and Automation Letters},
  year={2025},
  publisher={IEEE}
}

@article{pinto2017asymmetric,
  title={Asymmetric actor critic for image-based robot learning},
  author={Pinto, Lerrel and Andrychowicz, Marcin and Welinder, Peter and Zaremba, Wojciech and Abbeel, Pieter},
  journal={arXiv preprint arXiv:1710.06542},
  year={2017}
}

@article{lee2010control,
  title={Control of complex maneuvers for a quadrotor UAV using geometric methods on SE (3)},
  author={Lee, Taeyoung and Leok, Melvin and McClamroch, N Harris},
  journal={arXiv preprint arXiv:1003.2005},
  year={2010}
}

@article{faessler2017differential,
  title={Differential flatness of quadrotor dynamics subject to rotor drag for accurate tracking of high-speed trajectories},
  author={Faessler, Matthias and Franchi, Antonio and Scaramuzza, Davide},
  journal={IEEE Robotics and Automation Letters},
  volume={3},
  number={2},
  pages={620--626},
  year={2017},
  publisher={IEEE}
}

@article{zhang2024robust,
  title={Robust Locomotion Policy with Adaptive Lipschitz Constraint for Legged Robots},
  author={Zhang, Yang and Nie, Buqing and Gao, Yue},
  journal={IEEE Robotics and Automation Letters},
  year={2024},
  publisher={IEEE}
}

@article{schulman2017proximal,
  title={Proximal policy optimization algorithms},
  author={Schulman, John and Wolski, Filip and Dhariwal, Prafulla and Radford, Alec and Klimov, Oleg},
  journal={arXiv preprint arXiv:1707.06347},
  year={2017}
}

@ARTICLE{HowFast,
  author={Falanga, Davide and Kim, Suseong and Scaramuzza, Davide},
  journal={IEEE Robotics and Automation Letters}, 
  title={How Fast Is Too Fast? The Role of Perception Latency in High-Speed Sense and Avoid}, 
  year={2019},
  volume={4},
  number={2},
  pages={1884-1891},
  doi={10.1109/LRA.2019.2898117}}

@inproceedings{madaan2020airsim,
  title={Airsim drone racing lab},
  author={Madaan, Ratnesh and Gyde, Nicholas and Vemprala, Sai and Brown, Matthew and Nagami, Keiko and Taubner, Tim and Cristofalo, Eric and Scaramuzza, Davide and Schwager, Mac and Kapoor, Ashish},
  booktitle={Neurips 2019 competition and demonstration track},
  pages={177--191},
  year={2020},
  organization={PMLR}
}

@inproceedings{airsim2017fsr,
  author = {Shital Shah and Debadeepta Dey and Chris Lovett and Ashish Kapoor},
  title = {AirSim: High-Fidelity Visual and Physical Simulation for Autonomous Vehicles},
  year = {2017},
  booktitle = {Field and Service Robotics},
  eprint = {arXiv:1705.05065},
  url = {https://arxiv.org/abs/1705.05065}
}

@article{zhang2025learning,
  title={Learning vision-based agile flight via differentiable physics},
  author={Zhang, Yuang and Hu, Yu and Song, Yunlong and Zou, Danping and Lin, Weiyao},
  journal={Nature Machine Intelligence},
  pages={1--13},
  year={2025},
  publisher={Nature Publishing Group UK London}
}

@article{yin2025taco,
  title={TACO: General Acrobatic Flight Control via Target-and-Command-Oriented Reinforcement Learning},
  author={Yin, Zikang and Zheng, Canlun and Guo, Shiliang and Wang, Zhikun and Zhao, Shiyu},
  journal={arXiv preprint arXiv:2503.01125},
  year={2025}
}


 




\vfill

\end{document}